\documentclass[11pt,a4paper]{article}
\usepackage[hyperref]{acl2021}
\usepackage{times}
\usepackage{latexsym}

\usepackage{microtype}

\aclfinalcopy 
\def\aclpaperid{549} 

\usepackage{multirow}
\usepackage{amsmath}
\usepackage{capt-of}
\usepackage{tabularx}
\usepackage[caption=false]{subfig}
\usepackage{epsfig}
\usepackage{amssymb}
\usepackage{amsfonts}
\usepackage{booktabs}
\usepackage{scalerel}
\usepackage[inline]{enumitem}
\usepackage{listings}
\usepackage{varwidth}
\usepackage[export]{adjustbox}
\usepackage{tikz}
\usetikzlibrary{tikzmark}
\usepackage{cleveref}

\usepackage{stmaryrd}
\usepackage{bbm}

\newcommand{\tabincell}[2]{\begin{tabular}{@{}#1@{}}#2\end{tabular}}

\usepackage{algorithm}
\usepackage[noend]{algpseudocode}

\definecolor{deepblue}{rgb}{0,0,0.5}
\definecolor{officeblue}{RGB}{0,102,204}
\definecolor{deepred}{rgb}{0.6,0,0}
\definecolor{deepgreen}{rgb}{0,0.5,0}
\definecolor{mybrickred}{RGB}{182,50,28}

\definecolor{fillcolor}{RGB}{216,217,252}

\algnewcommand\algorithmicrequireb{{\hspace{0.85cm}}}
\algnewcommand\INPTDESCB{\item[\algorithmicrequireb]}

\algnewcommand\algorithmicfuncdesc{\textbf{Function:}}
\algnewcommand\FUNCDESC{\item[\algorithmicfuncdesc]}
\algnewcommand\algorithmicfuncdescb{{\hspace{1.48cm}}}
\algnewcommand\FUNCDESCB{\item[\algorithmicfuncdescb]}
\algnewcommand{\algorithmicgoto}{\textbf{goto}}
\algnewcommand{\Goto}[1]{\algorithmicgoto~\ref{#1}}

\usepackage{amsmath,amsfonts,bm}

\def\eqref#1{equation~\ref{#1}}

\def\1{\bm{1}}

\DeclareMathAlphabet{\mathsfit}{\encodingdefault}{\sfdefault}{m}{sl}
\SetMathAlphabet{\mathsfit}{bold}{\encodingdefault}{\sfdefault}{bx}{n}

\DeclareMathOperator*{\argmin}{arg\,min}

\newcommand*\samethanks[1][\value{footnote}]{\footnotemark[#1]}

\newcommand\ours{\textsc{xTune}}

    \makeatletter
\def\@fnsymbol#1{\ensuremath{\ifcase#1\or *\or \diamond\or
   \mathsection\or \mathparagraph\or \|\or **\or \dagger\dagger
   \or \ddagger\ddagger \else\@ctrerr\fi}}
    \makeatother

\title{Consistency Regularization for Cross-Lingual Fine-Tuning}

\author{Bo Zheng$^\dag$\thanks{\ \  Contribution during internship at Microsoft Research.},~~Li Dong$^\ddag$,~~Shaohan Huang$^\ddag$,~~Wenhui Wang$^{\ddag}$,~~Zewen Chi$^{\ddag}\samethanks[1]$\\
\textbf{Saksham Singhal}$^{\ddag}$\textbf{,}~~\textbf{Wanxiang Che}$^{\dag}  $\textbf{,}~~\textbf{Ting Liu}$^\dag$\textbf{,}~~\textbf{Xia Song}$^\ddag$\textbf{,}~~\textbf{Furu Wei}$^\ddag$\\
$^\dag$Harbin Institute of Technology \\
$^\ddag$Microsoft Corporation \\
\texttt{\{bzheng,car,tliu\}@ir.hit.edu.cn} \\
\texttt{\{lidong1,shaohanh,wenwan,saksingh,xiaso,fuwei\}@microsoft.com} \\}

\date{}

\begin{document}
\maketitle
\begin{abstract}
Fine-tuning pre-trained cross-lingual language models can transfer task-specific supervision from one language to the others. In this work, we propose to improve cross-lingual fine-tuning with consistency regularization. Specifically, we use \textit{example consistency} regularization to penalize the prediction sensitivity to four types of data augmentations, i.e., subword sampling, Gaussian noise, code-switch substitution, and machine translation. In addition, we employ \textit{model consistency} to regularize the models trained with two augmented versions of the same training set. Experimental results on the XTREME benchmark show that our method\footnote{The code is available at \url{https://github.com/bozheng-hit/xTune}.} significantly improves cross-lingual fine-tuning across various tasks, including text classification, question answering, and sequence labeling.
\end{abstract}

\section{Introduction}
\label{sec:intro}

Pre-trained cross-lingual language models~\citep{DBLP:conf/nips/ConneauL19, DBLP:conf/acl/ConneauKGCWGGOZ20, DBLP:journals/corr/abs-2007-07834} have shown great transferability across languages.
By fine-tuning on labeled data in a source language, the models can generalize to other target languages, even without any additional training.
Such generalization ability reduces the required annotation efforts, which is prohibitively expensive for low-resource languages.

Recent work has demonstrated that data augmentation is helpful for cross-lingual transfer, e.g., translating source language training data into target languages~\citep{DBLP:journals/corr/abs-1905-11471}, and generating code-switch data by randomly replacing input words in the source language with translated words in target languages~\citep{DBLP:conf/ijcai/QinN0C20}.
By populating the dataset, their fine-tuning still treats training instances independently, without considering the inherent correlations between the original input and its augmented example.
In contrast, we propose to utilize consistency regularization to better leverage data augmentation for cross-lingual fine-tuning.
Intuitively, for a semantic-preserving augmentation strategy, the predicted result of the original input should be similar to its augmented one.
For example, the classification predictions of an English sentence and its translation tend to remain consistent.

In this work, we introduce a cross-lingual fine-tuning method \ours{} that is enhanced by consistency regularization and data augmentation.
First, \textit{example consistency} regularization enforces the model predictions to be more consistent for semantic-preserving augmentations.
The regularizer penalizes the model sensitivity to different surface forms of the same example (e.g., texts written in different languages), which implicitly encourages cross-lingual transferability.
Second, we introduce \textit{model consistency} to regularize the models trained with various augmentation strategies.
Specifically, given two augmented versions of the same training set, we encourage the models trained on these two datasets to make consistent predictions for the same example.
The method enforces the corpus-level consistency between the distributions learned by two models.

Under the proposed fine-tuning framework, we study four strategies of data augmentation, i.e., subword sampling~\cite{DBLP:conf/acl/Kudo18}, code-switch substitution~\cite{DBLP:conf/ijcai/QinN0C20}, Gaussian noise~\cite{DBLP:journals/corr/abs-2008-03156}, and machine translation.
We evaluate \ours{} on the XTREME benchmark~\cite{DBLP:conf/icml/HuRSNFJ20}, including three different tasks on seven datasets.
Experimental results show that our method outperforms conventional fine-tuning with data augmentation.
We also demonstrate that \ours{} is flexible to be plugged in various tasks, such as classification, span extraction, and sequence labeling.

We summarize our contributions as follows:
\begin{itemize}
\item We propose \ours{}, a cross-lingual fine-tuning method to better utilize data augmentations based on consistency regularization.
\item We study four types of data augmentations that can be easily plugged into cross-lingual fine-tuning.
\item We give instructions on how to apply \ours{} to various downstream tasks, such as classification, span extraction, and sequence labeling.
\item We conduct extensive experiments to show that \ours{} consistently improves the performance of cross-lingual fine-tuning.
\end{itemize}

\section{Related Work}

\paragraph{Cross-Lingual Transfer}
Besides learning cross-lingual word embeddings~\citep{DBLP:journals/corr/MikolovLS13,DBLP:conf/eacl/FaruquiD14,DBLP:conf/acl/GuoCYWL15, DBLP:conf/emnlp/XuYOW18,DBLP:conf/emnlp/WangCGLL19}, most recent work of cross-lingual transfer is based on pre-trained cross-lingual language models~\citep{DBLP:conf/nips/ConneauL19,DBLP:conf/acl/ConneauKGCWGGOZ20,DBLP:journals/corr/abs-2007-07834}.
These models generate multilingual contextualized word representations for different languages with a shared encoder and show promising cross-lingual transferability.

\paragraph{Cross-Lingual Data Augmentation}
Machine translation has been successfully applied to the cross-lingual scenario as data augmentation.
A common way to use machine translation is to fine-tune models on both source language training data and translated data in all target languages.
Furthermore, \citet{DBLP:journals/corr/abs-1905-11471} proposed to replace a segment of source language input text with its translation in another language. 
However, it is usually impossible to map the labels in source language data into target language translations for token-level tasks.
\citet{DBLP:conf/emnlp/ZhangZF19} used code-mixing to perform the syntactic transfer in cross-lingual dependency parsing.
\citet{DBLP:conf/acl/FeiZJ20} constructed pseudo translated target corpora from the gold-standard annotations of the source languages for cross-lingual semantic role labeling. \citet{DBLP:journals/corr/abs-2009-05166} proposed an additional Kullback-Leibler divergence self-teaching loss for model training, based on auto-generated soft pseudo-labels for translated text in the target language.
Besides, \citet{DBLP:conf/ijcai/QinN0C20} fine-tuned models on multilingual code-switch data, which achieves considerable improvements.

\paragraph{Consistency Regularization}
One strand of work in consistency regularization focused on regularizing model predictions to be invariant to small perturbations on image data. 
The small perturbations can be random noise~\citep{stable:training}, adversarial noise~\citep{DBLP:journals/pami/MiyatoMKI19, DBLP:conf/nips/CarmonRSDL19} and various data augmentation approaches~\citep{DBLP:conf/icml/HuMTMS17, DBLP:conf/cvpr/YeZYC19,DBLP:conf/nips/XieDHL020}.
Similar ideas are used in the natural language processing area.
Both adversarial noise~\citep{DBLP:conf/iclr/ZhuCGSGL20,DBLP:conf/acl/JiangHCLGZ20,DBLP:journals/corr/abs-2004-08994} and sampled Gaussian noise~\citep{DBLP:journals/corr/abs-2008-03156} are adopted to augment input word embeddings.
Another strand of work focused on consistency under different model parameters~\citep{DBLP:conf/iclr/TarvainenV17,DBLP:conf/iclr/AthiwaratkunFIW19}, which is complementary to the first strand. 
We focus on the cross-lingual setting, where consistency regularization has not been fully explored.

\begin{figure*}[t]
\centering
\includegraphics[scale=0.45]{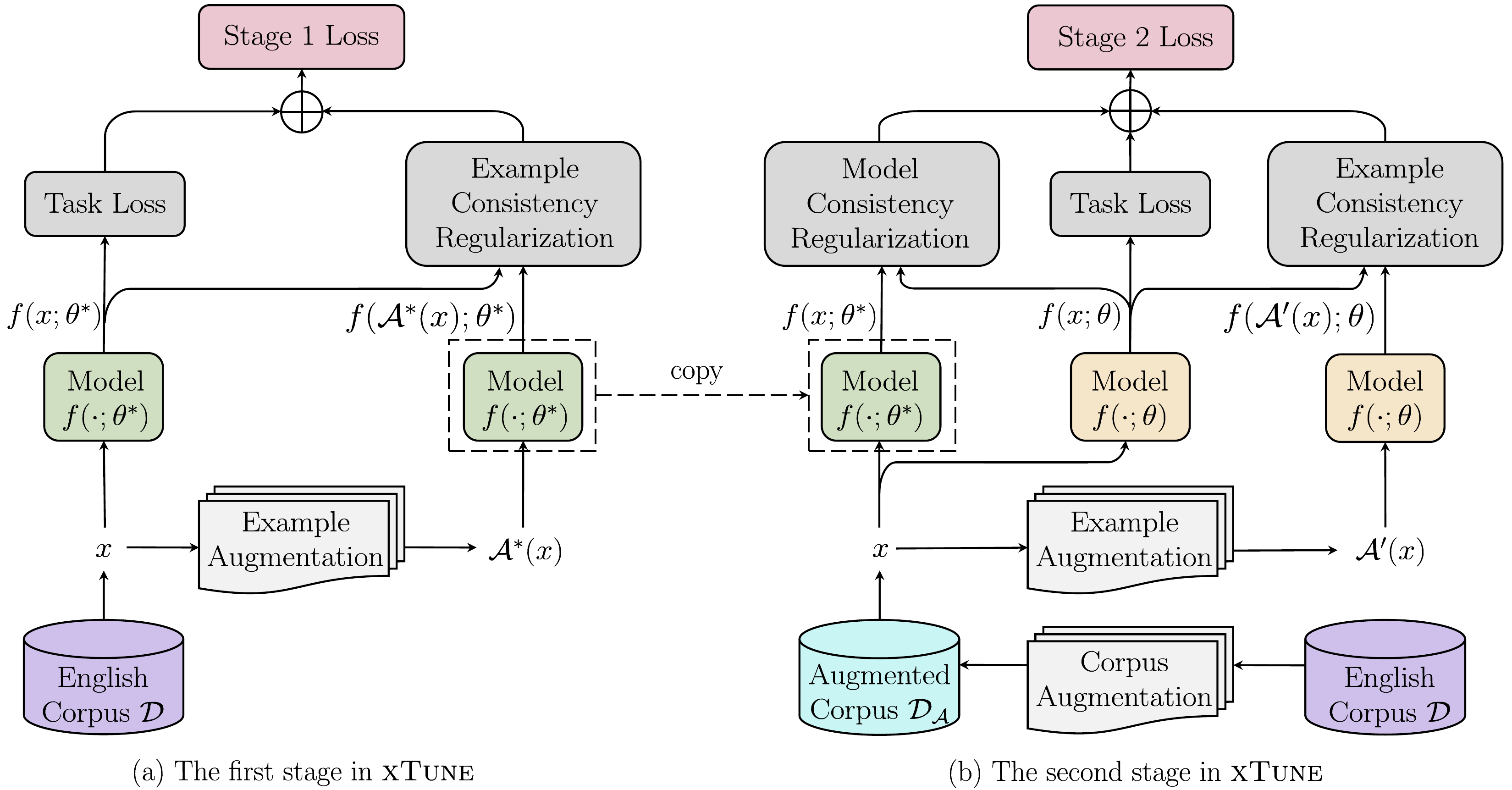}
\caption{Overview of our two-stage fine-tuning algorithm. The model parameters $f(\cdot;\theta^{*})$ in the second stage are copied from the first stage.}
\label{fig:stb-framework}
\end{figure*}

\section{Methods}
\label{sec:methods}

Conventional cross-lingual fine-tuning trains a pre-trained language model on the source language and directly evaluates it on other languages, which is also known as the setting of zero-shot cross-lingual fine-tuning.
Specifically, given a training corpus $\mathcal{D}$ in the source language (typically in English), and a model $f(\cdot; \theta)$ that predicts task-specific probability distributions, we define the loss of cross-lingual fine-tuning as:
\begin{equation*}
\label{eq:task-obj}
\mathcal{L}^\text{task}(\mathcal{D}, \theta) = \sum_{x \in \mathcal{D}} \ell(f(x;\theta), G(x)),
\end{equation*}
where $G(x)$ denotes the ground-truth label of example $x$, $\ell(\cdot,\cdot)$ is the loss function depending on the downstream task.

Apart from vanilla cross-lingual fine-tuning on the source language, recent work shows that data augmentation is helpful to improve performance on the target languages.
For example, \citet{DBLP:conf/nips/ConneauL19} add translated examples to the training set for better cross-lingual transfer.
Let $\mathcal{A}(\cdot)$ be a cross-lingual data augmentation strategy (such as code-switch substitution), and $\mathcal{D_{\mathcal{A}}}=\mathcal{D} \cup \{ \mathcal{A}(x) \mid x \in \mathcal{D} \}$ be the augmented training corpus, the fine-tuning loss is $\mathcal{L}^\text{task}(\mathcal{D_{A}}, \theta)$.
Notice that it is non-trivial to apply some augmentations for token-level tasks directly. For instance, in part-of-speech tagging, the labels of source language examples can not be mapped to the translated examples because of the lack of explicit alignments.

\subsection{\ours{}: Cross-Lingual Fine-Tuning with Consistency Regularization}
\label{sec:stability}

We propose to improve cross-lingual fine-tuning with two consistency regularization methods, so that we can effectively leverage cross-lingual data augmentations.

\subsubsection{Example Consistency Regularization}
\label{sec:stability:exp-reg}

In order to encourage consistent predictions for an example and its semantically equivalent augmentation, we introduce \textit{example consistency} regularization, which is defined as follows:
\begin{align}
\mathcal{R}_1(\mathcal{D}, \theta, \mathcal{A}) = &\sum_{x \in \mathcal{D}} \text{KL}_{\text{S}}(f({\color{blue}x};\theta){\parallel}f({\color{blue}\mathcal{A}(x)};\theta)), \nonumber \\
\text{KL}_{\text{S}}(P, Q) =~&\text{KL}(\mathrm{stopgrad}(P){\parallel}Q) + \nonumber \\
& \text{KL}(\mathrm{stopgrad}(Q){\parallel}P) \nonumber
\end{align}
where $\text{KL}_{\text{S}}(\cdot)$ is the symmertrical Kullback-Leibler divergence.
The regularizer encourages the predicted distributions $f(x;\theta)$ and $f(\mathcal{A}(x);\theta)$ to agree with each other.
The $\mathrm{stopgrad}(\cdot)$ operation\footnote{Implemented by \texttt{.detach()} in PyTorch.} is used to stop back-propagating gradients, which is also employed in~\citep{DBLP:conf/acl/JiangHCLGZ20, DBLP:journals/corr/abs-2004-08994}.
The ablation studies in Section~\ref{sec:results} empirically show that the operation improves fine-tuning performance.

\subsubsection{Model Consistency Regularization}
\label{sec:stability:cor-reg}

While the example consistency regularization is conducted at the example level, we propose the \textit{model consistency} to further regularize the model training at the corpus level.
The regularization is conducted at two stages.
First, we obtain a fine-tuned model $\theta^{*}$ on the training corpus $\mathcal{D}$:
\begin{align}
\theta^{*} = \mathop{\argmin}\limits_{\theta_1} \mathcal{L}^\text{task}(\mathcal{D}, \theta_1) . \nonumber
\end{align}
In the second stage, we keep the parameters $\theta^{*}$ fixed. The regularization term is defined as: 
\begin{align}
\mathcal{R}_2(\mathcal{D}_{\mathcal{A}}, \theta, \theta^{*}) = \sum_{x \in \mathcal{D}_{\mathcal{A}}} \text{KL}(f(x;\textcolor{blue}{\theta^{*}}){\parallel}f(x;\textcolor{blue}{\theta})) \nonumber
\end{align}
where $\mathcal{D_{A}}$ is the augmented training corpus, and $\text{KL}(\cdot)$ is Kullback-Leibler divergence.
For each example $x$ of the augmented training corpus $\mathcal{D_{A}}$, the model consistency regularization encourages the prediction $f(x;\theta)$ to be consistent with $f(x;\theta^{*})$.
The regularizer enforces the corpus-level consistency between the distributions learned by two models.

An unobvious advantage of model consistency regularization is the flexibility with respect to data augmentation strategies.
For the example of part-of-speech tagging, even though the labels can not be directly projected from an English sentence to its translation, we are still able to employ the regularizer.
Because the term $\mathcal{R}_2$ is put on the same example $x \in \mathcal{D_{A}}$, we can always align the token-level predictions of the models $\theta$ and $\theta^{*}$.

\subsubsection{Full \ours{} Fine-Tuning}
\label{sec:stability:stb-obj}

As shown in Figure~\ref{fig:stb-framework}, we combine example consistency regularization $\mathcal{R}_{1}$ and model consistency regularization $\mathcal{R}_{2}$ as a two-stage fine-tuning process.
Formally, we fine-tune a model with $\mathcal{R}_{1}$ in the first stage:
\begin{align} 
\theta^{*} = \mathop{\argmin}\limits_{\theta_1}  \mathcal{L}^\text{task}(\mathcal{D}, \theta_1) + \mathcal{R}_1(\mathcal{D}, \theta_1, \mathcal{A}^{*}) \nonumber
\end{align}
where the parameters $\theta^{*}$ are kept fixed for $\mathcal{R}_{2}$ in the second stage.
Then the final loss is computed via:
\begin{align} 
\mathcal{L}^{\textsc{xTune}} =~&\mathcal{L}^\text{task}(\mathcal{D}_\mathcal{A}, \theta) \nonumber \\
&+ \lambda_{1}\mathcal{R}_1(\mathcal{D}_\mathcal{A}, \theta, \mathcal{A}') \nonumber \\
&+ \lambda_{2}\mathcal{R}_2(\mathcal{D}_\mathcal{A}, \theta, \theta^{*}) \nonumber
\end{align}
where $\lambda_{1}$ and $\lambda_{2}$ are the corresponding weights of two regularization methods. 
Notice that the data augmentation strategies $\mathcal{A}$, $\mathcal{A}'$, and $\mathcal{A}^{*}$ can be either different or the same, which are tuned as hyper-parameters.

\begin{figure}[t]
\centering
\includegraphics[scale=0.485]{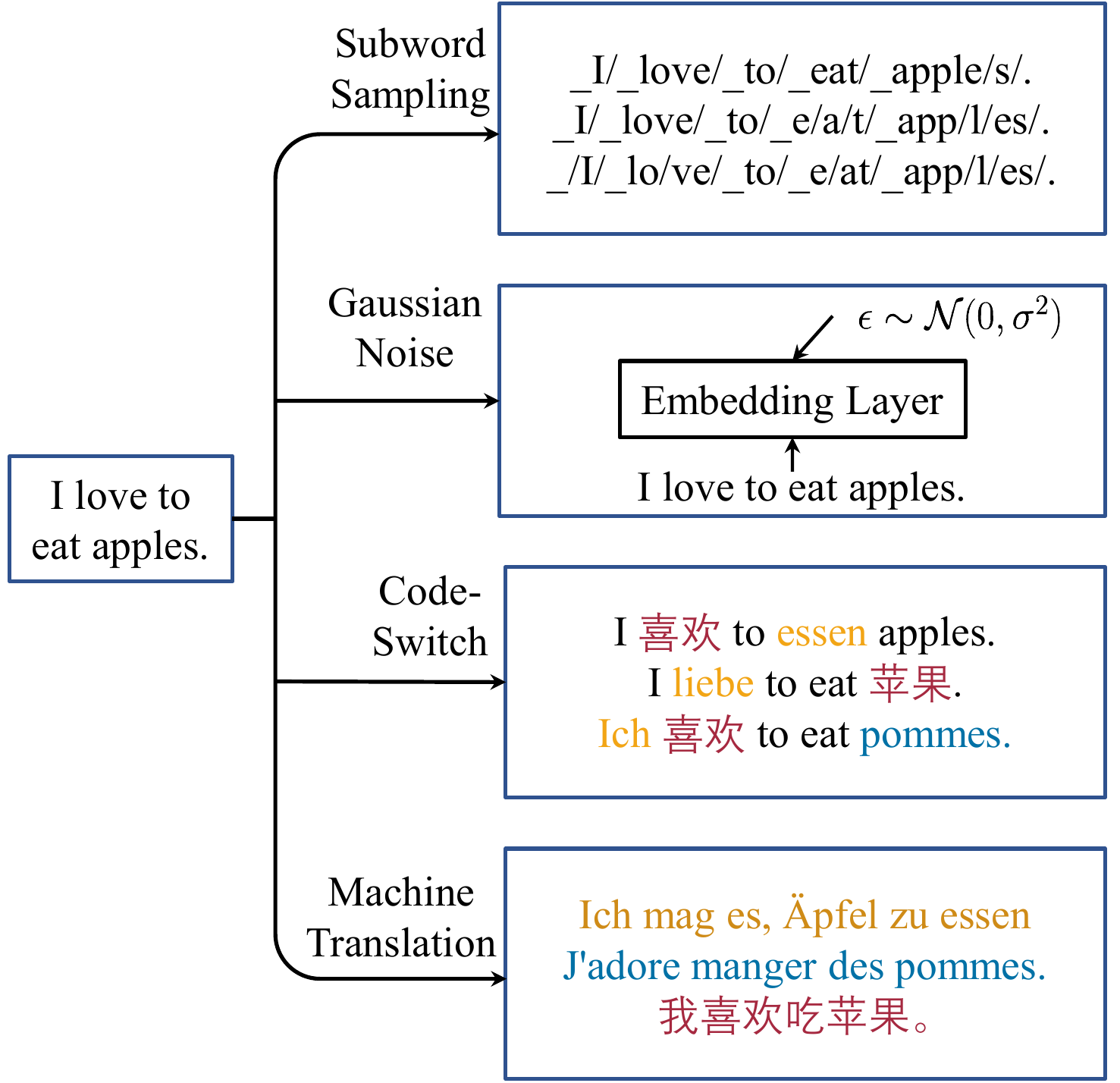}
\caption{Cross-lingual data augmentation strategies.}
\label{fig:data-aug}
\end{figure}

\subsection{Data Augmentation}
\label{sec:noise}
We consider four types of data augmentation strategies in this work, which are shown in Figure~\ref{fig:data-aug}. 
We aim to study the impact of different data augmentation strategies on cross-lingual transferability.
\subsubsection{Subword Sampling}
\label{sec:noise:subword}

Representing a sentence in different subword sequences can be viewed as a data augmentation strategy~\citep{DBLP:conf/acl/Kudo18, DBLP:conf/acl/ProvilkovEV20}. We utilize XLM-R~\citep{DBLP:conf/acl/ConneauKGCWGGOZ20} as our pre-trained cross-lingual language model, while it applies subword tokenization directly on raw text data using SentencePiece~\cite{DBLP:conf/emnlp/KudoR18} with a unigram language model~\cite{DBLP:conf/acl/Kudo18}.
As one of our data augmentation strategies, we apply the on-the-fly subword sampling algorithm in the unigram language model to generate multiple subword sequences.  

\subsubsection{Gaussian Noise}
\label{sec:noise:gaussian}
Most data augmentation strategies in NLP change input text discretely, while we directly add random perturbation noise sampled from Gaussian distribution on the input embedding layer to conduct data augmentation.  
When combining this data augmentation with example consistency $\mathcal{R}_{1}$, the method is similar to the stability training~\citep{stable:training}, random perturbation training~\citep{DBLP:journals/pami/MiyatoMKI19} and the R3F method~\citep{DBLP:journals/corr/abs-2008-03156}.
We also explore Gaussian noise's capability to generate new examples on continuous input space for conventional fine-tuning.

\subsubsection{Code-Switch Substitution}
\label{sec:noise:codeswitch}

Anchor points have been shown useful to improve cross-lingual transferability.~\citet{DBLP:conf/acl/ConneauWLZS20} analyzed the impact of anchor points in pre-training cross-lingual language models. 
Following~\citet{DBLP:conf/ijcai/QinN0C20}, we generate code-switch data in multiple languages as data augmentation. 
We randomly select words in the original text in the source language and replace them with target language words in the bilingual dictionaries to obtain code-switch data. Intuitively, this type of data augmentation explicitly helps pre-trained cross-lingual models align the multilingual vector space by the replaced anchor points.

\subsubsection{Machine Translation}
\label{sec:noise:mt}
Machine translation has been proved to be an effective data augmentation strategy~\citep{DBLP:journals/corr/abs-1905-11471} under the cross-lingual scenario.
However, the ground-truth labels of translated data can be unavailable for token-level tasks~(see Section~\ref{sec:methods}), which disables conventional fine-tuning on the augmented data. 
Meanwhile, our proposed model consistency $\mathcal{R}_2$ can not only serve as consistency regularization but also can be viewed as a self-training objective to enable semi-supervised training on the unlabeled target language translations.

\subsection{Task Adaptation}

We give instructions on how to apply \ours{} to various downstream tasks, i.e., classification, span extraction, and sequence labeling.
By default, we use model consistency $\mathcal{R}_2$ in full \ours{}.
We describe the usage of example consistency $\mathcal{R}_1$ as follows.

\subsubsection{Classification}
\label{sec:stability:class}

For classification task, the model is expected to predict one distribution per example on $n_\text{label}$ types, i.e., model $f(\cdot;\theta)$ should predict a probability distribution $p_\text{cls} \in \mathbb{R}^{n_\text{label}}$. Thus we can directly use example consistency $\mathcal{R}_{1}$ to regularize the consistency of the two distributions for all four types of our data augmentation strategies.

\subsubsection{Span Extraction}
\label{sec:stability:span}
For span extraction task, the model is expected to predict two distributions per example $p_\text{start}, p_\text{end} \in \mathbb{R}^{n_\text{subword}}$, indicating the probability distribution of where the answer span starts and ends, $n_\text{subword}$ denotes the length of the tokenized input text. 
For Gaussian noise, the subword sequence remains unchanged so that example consistency $\mathcal{R}_{1}$ can be directly applied to the two distributions.
Since subword sampling and code-switch substitution will change $n_\text{subword}$, we control the ratio of words to be modified and utilize example consistency $\mathcal{R}_{1}$ on unchanged positions only.
We do not use the example consistency $\mathcal{R}_{1}$ for machine translation because it is impossible to explicitly align the two distributions.

\subsubsection{Sequence Labeling}
\label{sec:stability:seq}
Recent pre-trained language models generate representations at the subword-level. For sequence labeling tasks, these models predict label distributions on each word's first subword. Therefore, the model is expected to predict $n_\text{word}$ probability distributions per example on $n_\text{label}$ types.
Unlike span extraction, subword sampling, code-switch substitution, and Gaussian noise do not change $n_\text{word}$. Thus the three data augmentation strategies will not affect the usage of example consistency $\mathcal{R}_{1}$. 
Although word alignment is a possible solution to map the predicted label distributions between translation pairs, the word alignment process will introduce more noise. Therefore, we do not employ machine translation as data augmentation for the example consistency $\mathcal{R}_{1}$.

\section{Experiments}
\label{sec:exp}

\subsection{Experiment Setup}
\label{sec:setup}

\begin{table*}[t]
\centering
\small
\setlength{\tabcolsep}{1.5mm}
\begin{tabular}{lcccccccc}
\toprule
\multirow{2}{*}{\bf Model} & \multicolumn{2}{c}{\bf Pair Sentence} & \multicolumn{2}{c}{\bf Structure Prediction} & \multicolumn{3}{c}{\bf Question Answering} &  \\
 & XNLI & PAWS-X & POS & NER & XQuAD & MLQA & TyDiQA \\ \midrule
\textbf{Metrics} & Acc. & Acc. & F1 & F1 & F1/EM & F1/EM & F1/EM & Avg.\\ \midrule
\multicolumn{9}{l}{\textit{Cross-lingual-transfer (models are fine-tuned on English training data without translation available)}} \\
\midrule
mBERT & 65.4 & 81.9 & 70.3 & 62.2 & 64.5/49.4 & 61.4/44.2 & 59.7/43.9 & 63.1  \\ 
XLM & 69.1 & 80.9 & 70.1 & 61.2 & 59.8/44.3 & 48.5/32.6 & 43.6/29.1 & 58.6 \\ 
X-STILTs~\citep{DBLP:journals/corr/abs-2005-13013} & 80.4 & 87.7 & 74.4 & 63.4 & 77.2/61.3 & 72.3/53.5 & 76.0/59.5 & 72.3 \\ 
VECO~\citep{veco} & 79.9 & 88.7 & 75.1 & 65.7 & 77.3/61.8 & 71.7/53.2 & 67.6/49.1 & 71.4 \\
$\text{XLM-R}_{\text{large}}$ & 79.2 & 86.4 & 72.6 & 65.4 & 76.6/60.8 & 71.6/53.2 & 65.1/45.0 & 70.0 \\ 
\ours{} & \textbf{82.6} & \textbf{89.8} & \textbf{78.5} & \textbf{69.3} & \textbf{79.4/64.4} & \textbf{74.4/56.2} & \textbf{74.8/59.4} & \textbf{74.9} \\ 
\midrule
\multicolumn{9}{l}{\textit{Translate-train-all (translation-based augmentation is available for English training data)}} \\
\midrule
VECO~\citep{veco} & 83.0 & 91.1 & 75.1 & 65.7 & 79.9/66.3 & 73.1/54.9 & 75.0/58.9 & 74.1 \\
FILTER~\citep{DBLP:journals/corr/abs-2009-05166} & 83.9 & 91.4 & 76.2 & 67.7 & 82.4/68.0 & \textbf{76.2/57.7} & 68.3/50.9 & 74.4 \\
$\text{XLM-R}_{\text{large}}$ & 82.6 & 90.4 & - & - & 80.2/65.9 & 72.8/54.3 & 66.5/47.7 & -\\
\ours{} & \textbf{84.8} & \textbf{91.6} & \textbf{79.3} & \textbf{69.9} & \textbf{82.5/69.0} & 75.0/57.1 & \textbf{75.4/60.8} & \textbf{76.5} \\
\bottomrule
\end{tabular}
\caption{Evaluation results on the XTREME benchmark.
Results of mBERT~\cite{DBLP:conf/naacl/DevlinCLT19}, XLM~\cite{DBLP:conf/nips/ConneauL19} and $\text{XLM-R}_{\text{large}}$~\cite{DBLP:conf/acl/ConneauKGCWGGOZ20} are taken from~\cite{DBLP:conf/icml/HuRSNFJ20}. Results of $\text{XLM-R}_{\text{large}}$ under the translate-train-all setting are from FILTER~\citep{DBLP:journals/corr/abs-2009-05166}.
The results of \ours{} are from the best models selected with the performance on the corresponding development set.
}
\label{table:benchmark-results}
\end{table*}

\begin{table*}[ht]
\centering
\small
\setlength{\tabcolsep}{1.5mm}
\begin{tabular}{lccccccc}
\toprule
\multirow{2}{*}{\bf Model} & \multicolumn{2}{c}{\bf Pair Sentence} & \multicolumn{2}{c}{\bf Structure Prediction} & \multicolumn{3}{c}{\bf Question Answering}  \\
 & XNLI & PAWS-X & POS & NER & XQuAD & MLQA & TyDiQA \\
 \midrule
\textbf{Metrics} & Acc. & Acc. & F1 & F1 & F1/EM & F1/EM & F1/EM\\ 
\midrule
\multicolumn{8}{l}{\textit{Cross-lingual-transfer (models are fine-tuned on English training data without translation available)}} \\
\midrule
$\text{XLM-R}_{\text{base}}$ & 74.9 & 84.9 & 75.6 & 61.8 & 71.9/56.4 & 65.0/47.1 & 55.4/38.3 \\ 
\ours{} & \textbf{77.7}  & \textbf{87.5} & \textbf{76.5} & \textbf{63.0} & \textbf{73.9/59.0} & \textbf{68.1/50.2} & \textbf{61.2/45.2} \\ 
~~with only \textit{example consistency} $\mathcal{R}_1$ & 77.6 & 87.2 & 76.3 & 62.4 & 73.6/58.6 & 67.6/49.7 &  60.7/44.4 \\ 
~~with only \textit{model consistency} $\mathcal{R}_2$ & 76.6 & 86.3 & 76.3 & \textbf{63.0} & 73.2/58.1 & 66.7/49.0 & 59.2/42.3 \\ 
\midrule
\multicolumn{8}{l}{\textit{Translate-train-all (translation-based augmentation is available for English training data)}} \\ 
\midrule
$\text{XLM-R}_{\text{base}}$ & 78.8 & 88.4 & - & - & 75.2/61.4 &  67.8/50.1 & 63.7/47.7\\
\ours{} & \textbf{80.6} & \textbf{89.4} & \textbf{77.8} & \textbf{63.7} & \textbf{78.1/64.4} & 69.7/\textbf{52.1}  & \textbf{65.9/51.1} \\ 
~~with only \textit{example consistency} $\mathcal{R}_1$  & 80.5 & 89.3 & - & - & 76.1/62.5 & 69.1/51.6 & 65.1/50.3 \\
~~with only \textit{model consistency} $\mathcal{R}_2$ & 78.9 & 88.5 & 76.6 & 63.5 & 77.4/63.4 & 68.7/51.1 & 64.5/48.7 \\ 
~~remove $\mathrm{stopgrad}$ in $\mathcal{R}_1$ & 80.2 & 89.1 & 76.8 & 63.4 & 77.3/63.4 & \textbf{69.9/52.1} & 65.1/50.5 \\ 
\bottomrule
\end{tabular}
\caption{
Ablation studies on the XTREME benchmark.
All numbers are averaged over five random seeds.
}
\label{table:base-results}
\end{table*}

\paragraph{Datasets} 
For our experiments, we select three types of cross-lingual understanding tasks from XTREME benchmark~\cite{DBLP:conf/icml/HuRSNFJ20}, including two classification datasets: XNLI~\cite{DBLP:conf/emnlp/ConneauRLWBSS18}, PAWS-X~\cite{DBLP:conf/emnlp/YangZTB19}, three span extraction datasets: XQuAD~\cite{DBLP:conf/acl/ArtetxeRY20}, MLQA~\cite{DBLP:conf/acl/LewisORRS20}, TyDiQA-GoldP~\cite{DBLP:journals/tacl/ClarkPNCGCK20}, and two sequence labeling datasets: NER~\cite{DBLP:conf/acl/PanZMNKJ17}, POS~\cite{d701bee1cabe492caf36340d6341e27b}. 
The statistics of the datasets are shown in the supplementary document.

\begin{table*}[ht]
\centering
\small
\setlength{\tabcolsep}{0.9mm}
\begin{tabular}{lcccccccccccccccc}
\toprule
{\bf Model} & {\bf en} & {\bf ar} & {\bf bg} & {\bf de} & {\bf el} & {\bf es} & {\bf fr} & {\bf hi} & {\bf ru} & {\bf sw} & {\bf th} & {\bf tr} & {\bf ur} & {\bf vi} & {\bf zh} & {\bf Avg.} \\
\midrule
\multicolumn{17}{l}{\textit{Cross-lingual-transfer (models are fine-tuned on English training data without translation available)}} \\
\midrule
R3F~\cite{DBLP:journals/corr/abs-2008-03156} & 89.4 & 80.6 & 84.6 & 83.7 & 83.6 & 85.1 & 84.2 & 77.3 & 82.3 & 72.6 & 79.4 & 80.7 & 74.2 & 81.1 & 80.1 & 81.2 \\ 
R4F~\cite{DBLP:journals/corr/abs-2008-03156} & \textbf{89.6} & 80.5 & 84.6 & 84.2 & 83.6 & 85.2 & 84.7 & 78.2 & 82.5 & 72.7 & 79.2 & 80.3 & 73.9 & 80.9 & 80.6 & 81.4 \\
$\text{XLM-R}_{\text{large}}$ & 88.7 & 77.2 & 83.0 & 82.5 & 80.8 & 83.7 & 82.2 & 75.6 & 79.1 & 71.2 & 77.4 & 78.0 & 71.7 & 79.3 & 78.2 & 79.2 \\
\ours{} & \textbf{89.6} & \textbf{81.6} & \textbf{85.9} & \textbf{84.8} & \textbf{84.3} & \textbf{86.5} & \textbf{85.4} & \textbf{80.5} & \textbf{82.8} & \textbf{73.3} & \textbf{80.3} & \textbf{82.1}& \textbf{77.1} & \textbf{83.0} & \textbf{82.3} & \textbf{82.6} \\
\midrule
\multicolumn{17}{l}{\textit{Translate-train-all (translation-based augmentation is available for English training data)}} \\
\midrule
FILTER~\citep{DBLP:journals/corr/abs-2009-05166} & 89.5 & 83.6 & 86.4 & 85.6 & 85.4 & 86.6 & 85.7 & 81.1 & 83.7 & 78.7 & 81.7 & 83.2 & 79.1 & 83.9 & 83.8 & 83.9 \\
$\text{XLM-R}_{\text{large}}$ & 88.6 & 82.2 & 85.2 & 84.5 & 84.5 & 85.7 & 84.2 & 80.8 & 81.8 & 77.0 & 80.2 & 82.1 & 77.7 & 82.6 & 82.7 & 82.6 \\
\ours{} & \textbf{89.9} & \textbf{84.0} & \textbf{87.0} & \textbf{86.5} & \textbf{86.2} & \textbf{87.4} & \textbf{86.6} & \textbf{83.2} & \textbf{85.2} & \textbf{80.0} & \textbf{82.7} & \textbf{84.1} & \textbf{79.6} & \textbf{84.8} & \textbf{84.3} & \textbf{84.8} \\
\bottomrule
\end{tabular}
\caption{XNLI accuracy scores for each language.
$\text{XLM-R}_{\text{large}}$ under the cross-lingual transfer setting are from~\cite{DBLP:conf/icml/HuRSNFJ20}. Results of $\text{XLM-R}_{\text{large}}$ under the translate-train-all setting are from~\citep{DBLP:journals/corr/abs-2009-05166}.}
\label{table:xnli-results}
\end{table*}

\begin{table}[t]
\centering
\small
\begin{tabular}{llccc}
\toprule
\textbf{Method} & \textbf{Model} & \textbf{XNLI} & \textbf{POS} & \textbf{MLQA}\\ \midrule
Baseline & $\text{XLM-R}_{\text{base}}$ & 74.9 & 75.6 & 65.0/47.1\\ \midrule
\multirow{3}{*}{\begin{tabular}[c]{@{}l@{}}Subword\\ Sampling\end{tabular}} &  Data Aug. & 75.3 & 75.8 & 64.7/46.7 \\ 
& \ours{}$_{\mathcal{R}_1}$ & {76.5} & {76.3} & {67.4/49.5} \\ 
& \ours{}$_{\mathcal{R}_2}$ & 75.8 & {76.3} & 66.7/49.0 \\ \midrule 
\multirow{3}{*}{\begin{tabular}[c]{@{}l@{}}Gaussian\\ Noise\end{tabular}} &  Data Aug. & 74.7 & 75.6 & 64.2/46.1 \\ 
& \ours{}$_{\mathcal{R}_1}$ & {76.3} & 75.7 & {66.7/48.9} \\ 
& \ours{}$_{\mathcal{R}_2}$ & 75.5 & {76.2} & 66.3/48.5 \\ \midrule
\multirow{3}{*}{\begin{tabular}[c]{@{}l@{}}Code-\\ Switch\end{tabular}} &  Data Aug. & 76.5 & 75.1 & 63.8/45.9 \\ 
& \ours{}$_{\mathcal{R}_1}$ & {77.6} & 75.8 & {67.6/49.7} \\ 
& \ours{}$_{\mathcal{R}_2}$ & 76.8 & {76.1} & 66.3/48.6 \\\midrule
\multirow{3}{*}{\begin{tabular}[c]{@{}l@{}}Machine\\ Translation\end{tabular}}&  Data Aug. & 78.8 & - & 67.8/50.1 \\
& \ours{}$_{\mathcal{R}_1}$ & {\textbf{79.7}} & - & - \\ 
& \ours{}$_{\mathcal{R}_2}$ & 78.9 & {\textbf{76.6}} & {\textbf{68.7/51.1}} \\
\bottomrule
\end{tabular}
\caption{Comparison between different data augmentation strategies.
``Data Aug.'' uses data augmentation for conventional fine-tuning.
``\ours{}$_{\mathcal{R}_1}$'' denotes fine-tuning with only example consistency $\mathcal{R}_{1}$. ``\ours{}$_{\mathcal{R}_2}$'' denotes fine-tuning with only model consistency $\mathcal{R}_{2}$.
}
\label{table:reg-results}
\end{table}

\paragraph{Fine-Tuning Settings}
We consider two typical fine-tuning settings from~\citet{DBLP:conf/acl/ConneauKGCWGGOZ20} and~\citet{DBLP:conf/icml/HuRSNFJ20} in our experiments, which are (1) \textit{cross-lingual transfer}: the models are fine-tuned on English training data without translation available, and directly evaluated on different target languages;
(2) \textit{translate-train-all}: translation-based augmentation is available, and the models are fine-tuned on the concatenation of English training data and its translated data on all target languages.
Since the official XTREME repository\footnote{\url{github.com/google-research/xtreme}} does not provide translated target language data for POS and NER, we use Google Translate to obtain translations for these two datasets. 

\paragraph{Implementation Details}
We utilize XLM-R~\citep{DBLP:conf/acl/ConneauKGCWGGOZ20} as our pre-trained cross-lingual language model. The bilingual dictionaries we used for code-switch substitution are from MUSE~\citep{lample2017unsupervised}.\footnote{\url{github.com/facebookresearch/MUSE}} 
For languages that cannot be found in MUSE, we ignore these languages since other bilingual dictionaries might be of poorer quality. 
For the POS dataset, we use the average-pooling strategy on subwords to obtain word representation since part-of-speech is related to different parts of words, depending on the language. 
We tune the hyper-parameter and select the model with the best average results over all the languages' development set. 
There are two datasets without development set in multi-languages.
For XQuAD, we tune the hyper-parameters with the development set of MLQA since they share the same training set and have a higher degree of overlap in languages. For TyDiQA-GoldP, we use the English test set as the development set. 
In order to make a fair comparison, the ratio of data augmentation in $\mathcal{D_{A}}$ is all set to 1.0.
The detailed hyper-parameters are shown in the supplementary document.

\subsection{Results}
\label{sec:results}
Table~\ref{table:benchmark-results} shows our results on XTREME. 
For the cross-lingual transfer setting, we outperform previous works on all seven cross-lingual language understanding datasets.\footnote{X-STILTs~\citep{DBLP:journals/corr/abs-2005-13013} uses additional SQuAD
v1.1 English training data for the TyDiQA-GoldP dataset, while we prefer a cleaner setting here.} 
Compared to $\text{XLM-R}_{\text{large}}$ baseline, we achieve an absolute 4.9-point improvement (70.0 vs. 74.9) on average over seven datasets. 
For the translate-train-all setting, we achieved state-of-the-art results on six of the seven datasets. 
Compared to FILTER,\footnote{FILTER directly selects the best model on the test set of XQuAD and TyDiQA-GoldP. Under this setting, we can obtain 83.1/69.7 for XQuAD, 75.5/61.1 for TyDiQA-GoldP.} we achieve an absolute 2.1-point improvement (74.4 vs. 76.5), and we do not need English translations during inference.

Table~\ref{table:base-results} shows how the two regularization methods affect the model performance separately. 
For the cross-lingual transfer setting, \ours{} achieves an absolute 2.8-point improvement compared to our implemented $\text{XLM-R}_{\text{base}}$ baseline. 
Meanwhile, fine-tuning with only example consistency $\mathcal{R}_{1}$ and model consistency $\mathcal{R}_{2}$ degrades the averaged results by 0.4 and 1.0 points, respectively.

For the translate-train-all setting, our proposed model consistency $\mathcal{R}_{2}$ enables training on POS and NER even if labels of target language translations are unavailable in these two datasets.
To make a fair comparison in the translate-train-all setting, we augment the English training corpus with target language translations when fine-tuning with only example consistency $\mathcal{R}_{1}$. 
Otherwise, we only use the English training corpus in the first stage, as shown in Figure~\ref{fig:stb-framework}(a).
Compared to \ours{}, the performance drop on two classification datasets under this setting is relatively small since $\mathcal{R}_{1}$ can be directly applied between translation-pairs in any languages. 
However, the performance is significantly degraded in three question answering datasets, where we can not align the predicted distributions between translation-pairs in $\mathcal{R}_{1}$. We use subword sampling as the data augmentation strategy in $\mathcal{R}_{1}$ for this situation.
Fine-tuning with only model consistency $\mathcal{R}_{2}$ degrades the overall performance by 1.1 points.
These results demonstrate that the two consistency regularization methods complement each other.
Besides, we observe that removing $\mathrm{stopgrad}$ degrades the overall performance by 0.5 points.

\begin{table}[t]
\centering
\small
\setlength{\tabcolsep}{1.3mm}
\begin{tabular}{lcc}
\toprule
\textbf{Model} & \textbf{Tatoeba} & \textbf{BUCC} \\ \midrule
$\text{XLM-R}_{\text{base}}$ (\textit{cross-lingual transfer}) & 74.2 & 78.2\\
$\text{XLM-R}_{\text{base}}$ (\textit{translate-train-all}) & 79.7 & 79.7 \\
\ours{} (\textit{translate-train-all}) & {\textbf{82.3}} & {\textbf{82.2}} \\
~~with only \textit{example consistency} $\mathcal{R}_1$ & 82.0 & 82.1 \\ 
~~with only \textit{model consistency} $\mathcal{R}_2$ & 79.5 & 79.0 \\
\bottomrule
\end{tabular}
\caption{Results of cross-lingual retrieval with the models fine-tuned on XNLI.}
\label{table:retrieval-results}
\end{table}

\begin{figure*}[t]
\centering
\includegraphics[width=\linewidth]{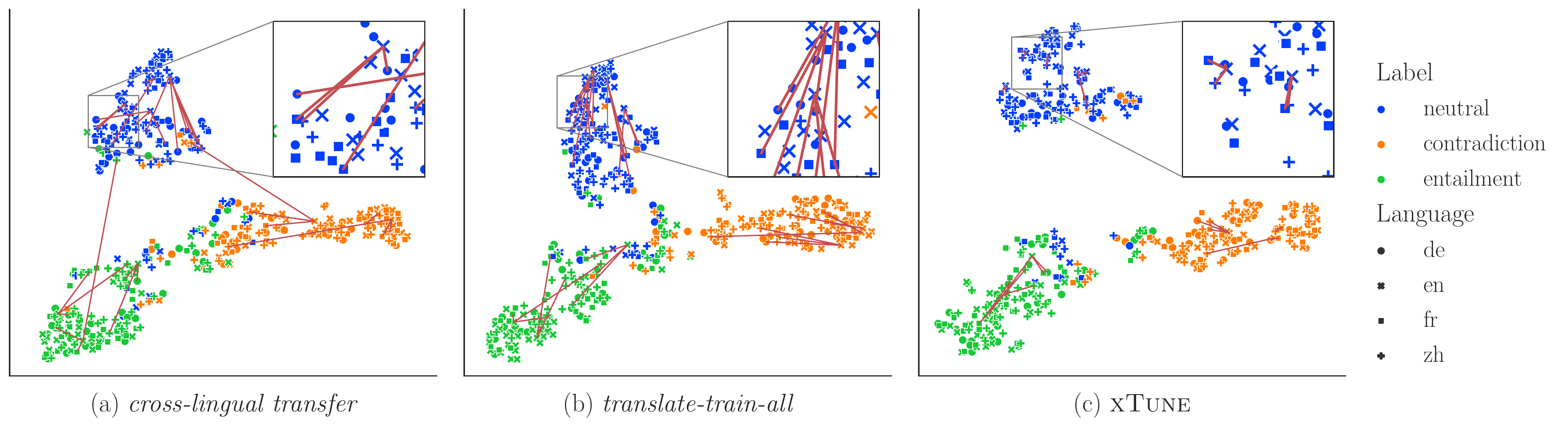}
\caption{t-SNE visualization of 100 examples in four languages from the XNLI development set (best viewed in color). We fine-tune the $\text{XLM-R}_\text{base}$ model on XNLI and use the hidden states of \texttt{[CLS]} symbol in the last layer. Examples with different labels are represented with different colors. Examples in different languages are represented with different markers. The red lines connect English examples and their translations in target languages.}
\label{fig:xnli}
\end{figure*}

Table~\ref{table:xnli-results} provides results of each language on the XNLI dataset.
For the cross-lingual transfer setting, we utilize code-switch substitution as data augmentation for both example consistency $\mathcal{R}_{1}$ and model consistency $\mathcal{R}_{2}$. We utilize all the bilingual dictionaries, except for English to Swahili and English to Urdu, which MUSE does not provide. 
Results show that our method outperforms all baselines on each language, even on Swahili (+2.2 points) and Urdu (+5.4 points), indicating our method can be generalized to low-resource languages even without corresponding machine translation systems or bilingual dictionaries.
For translate-train-all setting, we utilize machine translation as data augmentation for both example consistency $\mathcal{R}_{1}$ and model consistency $\mathcal{R}_{2}$. We improve the $\text{XLM-R}_{\text{large}}$ baseline by +2.2 points on average, while we still have +0.9 points on average compared to FILTER. It is worth mentioning that we do not need corresponding English translations during inference. 
Complete results on other datasets are provided in the supplementary document.

\subsection{Analysis}

\paragraph{It is better to employ data augmentation for consistency regularization than for conventional fine-tuning.}
As shown in Table~\ref{table:reg-results}, compared to employing data augmentation for conventional fine-tuning~(Data Aug.), our regularization methods~(\ours{}$_{\mathcal{R}_1}$, \ours{}$_{\mathcal{R}_2}$) consistently improve the model performance under all four data augmentation strategies. 
Since there is no labeled data on translations in POS and the issue of distribution alignment in example consistency $\mathcal{R}_{1}$, when machine translation is utilized as data augmentation, the results for Data Aug. and \ours{}$_{\mathcal{R}_1}$ in POS, as well as \ours{}$_{\mathcal{R}_1}$ in MLQA, are unavailable. 
We observe that Data Aug. can enhance the overall performance for coarse-grained tasks like XNLI, while our methods can further improve the results.
However, Data Aug. even causes the performance to degrade for fine-grained tasks like MLQA and POS.
In contrast, our proposed two consistency regularization methods improve the performance by a large margin~(e.g., for MLQA under code-switch data augmentation, Data Aug. decreases baseline by 1.2 points, while \ours{}$_{\mathcal{R}_1}$ increases baseline by 2.6 points).
We give detailed instructions on how to choose data augmentation strategies for \ours{} in the supplementary document.

\paragraph{\ours{} improves cross-lingual retrieval.}
We fine-tune the models on XNLI with different settings and compare their performance on two cross-lingual retrieval datasets.
Following~\citet{DBLP:journals/corr/abs-2007-07834} and~\citet{ DBLP:conf/icml/HuRSNFJ20}, we utilize representations averaged with hidden-states on the layer 8 of $\text{XLM-R}_{\text{base}}$.
As shown in Table~\ref{table:retrieval-results}, we observe significant improvement from the translate-train-all baseline to fine-tuning with only example consistency $\mathcal{R}_1$, this suggests regularizing the task-specific output of translation-pairs to be consistent also encourages the model to generate language-invariant representations. \ours{} only slightly improves upon this setting, indicating $\mathcal{R}_{1}$ between translation-pairs is the most important factor to improve cross-lingual retrieval task.

\paragraph{
\ours{} improves decision boundaries as well as the ability to generate language-invariant representations.
} As shown in Figure~\ref{fig:xnli}, we present t-SNE visualization of examples from the XNLI development set under three different settings. We observe the model fine-tuned with \ours{} significantly improves the decision boundaries of different labels. Besides, for an English example and its translations in other languages, the model fine-tuned with \ours{} generates more similar representations compared to the two baseline models. This observation is also consistent with the cross-lingual retrieval results in Table~\ref{table:retrieval-results}.

\section{Conclusion}
In this work, we present a cross-lingual fine-tuning framework \ours{} to make better use of data augmentation. 
We propose two consistency regularization methods that encourage the model to make consistent predictions for an example and its semantically equivalent data augmentation. 
We explore four types of cross-lingual data augmentation strategies.
We show that both example and model consistency regularization considerably boost the performance compared to directly fine-tuning on data augmentations. Meanwhile, model consistency regularization enables semi-supervised training on the unlabeled target language translations.
\ours{} combines the two regularization methods, and the experiments show that it can improve the performance by a large margin on the XTREME benchmark.

\section*{Acknowledgments}

Che is the corresponding author. This work was supported by the National Key R\&D Program of China via grant 2020AAA0106501 and the National Natural Science Foundation of China (NSFC) via grant 61976072 and 61772153.

\bibliography{xtune,anthology}
\bibliographystyle{acl_natbib}

\appendix
\section*{Appendix}
\label{sec:appendix}

\section{Statistics of XTREME Datasets}
\begin{table}[h]
\centering
\begin{tabular}{llll}
\toprule
{\bf Task}                                                  & {\bf Dataset} & $| \textbf{Train} |$ & $| \textbf{Lang} |$ \\ \midrule
\multirow{2}{*}{Classification}                       & XNLI    & 392K                  & 15                      \\
& PAWS-X  & 49.4K                 & 7                       \\ \midrule
Structured                                            & POS     & 21K                   & 33                      \\
Prediction                                            & NER     & 20K                   & 40                      \\ \midrule
\multirow{3}{*}{\tabincell{c}{Question \\ Answering}} & XQuAD   & 87K                   & 11                      \\
& MLQA    & 87K                   & 7                       \\
& TyDiQA  & 3.7K                  & 9                       \\ \bottomrule
\end{tabular}
\caption{Statistics for the datasets in the XTREME benchmark. we report the number of training examples ($| \text{Train} |$), and the number of languages ($| \text{Lang} |$).}
\label{table:dataset}
\end{table}

\section{Hyper-Parameters}

\begin{table*}[t]
\begin{tabular}{lcccccccc}
\toprule
                         & {\bf Variable}          & {\bf XNLI} & {\bf PAWS-X} & {\bf POS} & {\bf NER} & {\bf XQuAD} & {\bf MLQA} & {\bf TyDiQA} \\ \midrule
Stage 1                  & $\mathcal{A}^{*}$   & CS   & CS     & SS  & SS  & CS    & CS   & SS           \\ \midrule
\multirow{2}{*}{Stage 2} & $\mathcal{A}$    & CS   & CS     & SS  & SS  & SS    & SS   & SS           \\
                         & $\mathcal{A}'$       & CS   & CS     & SS  & SS  & SS    & SS   & SS           \\ \midrule
\multirow{2}{*}{Hyper-parameters}         & $\lambda_{1}$ & 5.0  & 5.0    & 5.0 & 5.0 & 5.0   & 5.0  & 5.0    \\ 
& $\lambda_{2}$ & 5.0  & 2.0    & 0.3 & 5.0 & 5.0  & 5.0  & 5.0 \\ 
\bottomrule
\end{tabular}
\caption{The best hyper-parameters used for \ours{} under the cross-lingual transfer setting. ``SS'', ``CS'', ``MT'' denote the data augmentation methods: subword sampling, code-switch substitution, and machine translation, respectively.}
\label{table:param-cross}
\end{table*}

\begin{table*}[t]
\begin{tabular}{lcccccccc}
\toprule
                         & {\bf Variable}          & {\bf XNLI} & {\bf PAWS-X} & {\bf POS} & {\bf NER} & {\bf XQuAD} & {\bf MLQA} & {\bf TyDiQA} \\ \midrule
Stage 1                  & $\mathcal{A}^{*}$   & MT   & MT     & SS  & SS  & CS    & CS   & SS           \\ \midrule
\multirow{2}{*}{Stage 2} & $\mathcal{A}$    & MT   & MT     & MT  & MT  & MT    & MT   & MT           \\
                         & $\mathcal{A}'$        & MT   & MT     & SS  & SS  & SS    & SS   & SS           \\ \midrule
\multirow{2}{*}{Hyper-parameters}         & $\lambda_{1}$ & 5.0  & 5.0    & 5.0 & 5.0 & 5.0   & 5.0  & 5.0    \\ 
& $\lambda_{2}$ & 1.0  & 1.0    & 0.3 & 1.0 & 0.1  & 0.5  & 0.3    \\ 
\bottomrule
\end{tabular}
\caption{The best hyper-parameters used for \ours{} under the translate-train-all setting. ``SS'', ``CS'', ``MT'' denote the data augmentation methods subword sampling, code-switch substitution, and machine translation, respectively.}
\label{table:param-translate}
\end{table*}

\begin{table}[t]
\centering
\small
\setlength{\tabcolsep}{1.2mm}
\begin{tabular}{lccccc}
\toprule
{\bf Method} & {\bf Model} & {\bf XNLI} & {\bf POS} & {\bf MLQA} & {\bf Avg.}\\ \midrule
- & $\text{XLM-R}_{\text{base}}$ & 10.6 & 20.8 & 20.3 & 17.2 \\ \midrule
\multirow{3}{*}{\begin{tabular}[c]{@{}l@{}}Subword \\ Sampling\end{tabular}} & Data Aug. & 10.5 & 20.5 & 20.2 & 17.1 \\ 
& \ours{}$_{\mathcal{R}_1}$ & {10.2} & 20.2 &  {19.6} & {16.7} \\ 
& \ours{}$_{\mathcal{R}_2}$ & 10.6 & {\textbf{20.1}} &  {19.8} & 16.8 \\ \midrule
\multirow{3}{*}{\begin{tabular}[c]{@{}l@{}}Gaussian \\ Noise\end{tabular}}& Data Aug. & 10.8 & 20.6 & 19.8 & 17.1 \\
& \ours{}$_{\mathcal{R}_1}$ & {10.5} & 20.7 & 19.8 & 17.0 \\ 
& \ours{}$_{\mathcal{R}_2}$ & 10.8 & {20.2} & 19.7 & {16.9} \\ \midrule 
\multirow{3}{*}{\begin{tabular}[c]{@{}l@{}}Code- \\ Switch\end{tabular}}& Data Aug. &  9.2 & 21.1 & 20.5 & 16.9 \\
& \ours{}$_{\mathcal{R}_1}$ & 9.1 & 20.7 & \textbf{19.4} & 16.4 \\ 
& \ours{}$_{\mathcal{R}_2}$ & {\textbf{8.8}} & {20.2} & 20.0 & {\textbf{16.3}} \\ \midrule
\multirow{3}{*}{\begin{tabular}[c]{@{}l@{}}Machine \\ Translation\end{tabular}}& Data Aug. & 7.2 & - & 17.9 & - \\
& \ours{}$_{\mathcal{R}_1}$ & {\textbf{6.9}} & - & - & - \\ 
& \ours{}$_{\mathcal{R}_2}$ & 7.2 & {\textbf{19.6}} & {\textbf{17.1}} & {\textbf{14.6}} \\ 
\bottomrule
\end{tabular}
\caption{Cross-lingual transfer gap, i.e., averaged performance drop between English and other languages in zero-shot transfer. A smaller gap indicates better transferability. For MLQA, we report the average of F1-scores and exact match scores.}
\label{table:transfer-gap}
\end{table}

For XNLI, PAWS-X, POS and NER, we fine-tune 10 epochs. For XQuAD and MLQA, we fine-tune 4 epochs. For TyDiQA-GoldP, we fine-tune 20 epochs and 10 epochs for base and large model, respectively. 
We select $\lambda_{1}$ in [1.0, 2.0, 5.0], $\lambda_{2}$ in [0.3, 0.5, 1.0, 2.0, 5.0]. 
For learning rate, we select in [5e-6, 7e-6, 1e-5, 1.5e-5] for large models, [7e-6, 1e-5, 2e-5, 3e-5] for base models. 
We use batch size 32 for all datasets and 10\% of total training steps for warmup with a linear learning rate schedule. Our experiments are conducted with a single 32GB Nvidia V100 GPU, and we use gradient accumulation for large-size models.
The other hyper-parameters for the two-stage \ours{} training are shown in Table~\ref{table:param-cross} and Table~\ref{table:param-translate}. 

\section{Results for Each Dataset and Language}
We provide detailed results for each dataset and language below. We compare our method against $\text{XLM-R}_\text{large}$ for cross-lingual transfer setting, FILTER~\citep{DBLP:journals/corr/abs-2009-05166} for translate-train-all setting.

\section{How to Select Data Augmentation Strategies in \ours{}}
We give instructions on selecting a proper data augmentation strategy depending on the corresponding task.
\subsection{Classification}
The two distribution in example consistency $\mathcal{R}_{1}$ can always be aligned. 
Therefore, we recommend using machine translation as data augmentation if the machine translation systems are available. 
Otherwise, the priority of our data augmentation strategies is code-switch substitution, subword sampling and Gaussian noise.
\subsection{Span Extraction}
The two distribution in example consistency $\mathcal{R}_{1}$ can not be aligned in translation-pairs. 
Therefore, it is impossible to use machine translation as data augmentation in example consistency $\mathcal{R}_{1}$. 
We prefer to use code-switch when applying example consistency $\mathcal{R}_{1}$ individually. 
However, when the training corpus is augmented with translations, since the bilingual dictionaries between arbitrary language pairs may not be available, we recommend using subword sampling in example consistency $\mathcal{R}_1$.
\subsection{Sequence Labeling}
Similar to span extraction, the two distribution in example consistency $\mathcal{R}_{1}$ can not be aligned in translation-pairs.
Therefore, we do not use machine translation in example consistency $\mathcal{R}_{1}$. 
Unlike classification and span extraction, sequence labeling requires finer-grained information and is more sensitive to noise. 
We found code-switch is worse than subword sampling as data augmentation in both example consistency $\mathcal{R}_{1}$ and model consistency $\mathcal{R}_{2}$, it will even degrade performance for certain hyper-parameters. 
Thus we recommend using subword sampling in example consistency $\mathcal{R}_{1}$, and use machine translation to augment the English training corpus if machine translation systems are available, otherwise subword sampling.

\section{Cross-Lingual Transfer Gap}
As shown in Table~\ref{table:transfer-gap}, the cross-lingual transfer gap can be reduced under all four data augmentation strategies. 
Meanwhile, we observe machine translation and code-switch substitution achieve a smaller cross-lingual transfer gap than the other two data augmentation methods. 
This suggests the data augmentation methods with cross-lingual knowledge have a greater improvement in cross-lingual transferability.
Although code-switch significantly reduces the transfer gap on XNLI, the improvement is relatively small on POS and MLQA under the cross-lingual transfer setting, indicating the noisy code-switch substitution will harm the cross-lingual transferability on finer-grained tasks.

\begin{table*}[ht]
\centering
\small
\setlength{\tabcolsep}{3.5mm}
\begin{tabular}{lcccccccc}
\toprule
{\bf Model} & {\bf en} & {\bf de} & {\bf es} & {\bf fr} & {\bf ja} & {\bf ko} & {\bf zh} & {\bf Avg.} \\ \midrule
\multicolumn{9}{l}{\textit{Cross-lingual-transfer (models are fine-tuned on English training data without translation available)}} \\ \midrule
$\text{XLM-R}_{\text{large}}$ & 94.7 & 89.7 & 90.1 & 90.4 & 78.7 & 79.0 & 82.3 & 86.4 \\
\ours{} & 96.0 & 92.5 & 92.2 & 92.7 & 84.9 & 84.2 & 86.6 & 89.8  \\
\midrule
\multicolumn{9}{l}{\textit{Translate-train-all (translation-based augmentation is available for English training data)}} \\ \midrule
FILTER~\citep{DBLP:journals/corr/abs-2009-05166} & 95.9 & 92.8 & 93.0 & 93.7 & 87.4 & 87.6 & 89.6 & 91.5 \\
\ours{} & 96.1 & 92.6 & 93.1 & 93.9 & 87.8 & 89.0 & 88.8 & 91.6 \\
\bottomrule
\end{tabular}
\caption{PAWSX results (accuracy scores) for each language. }
\label{table:pawsx-results}
\end{table*}

\begin{table*}[ht]
\centering
\scriptsize
\setlength{\tabcolsep}{1mm}
\begin{tabular}{lcccccccccccc}
\toprule
{\bf Model} & {\bf en} & {\bf ar} & {\bf de} & {\bf el} & {\bf es} & {\bf hi} & {\bf ru} & {\bf th} & {\bf tr} & {\bf vi} & {\bf zh} & {\bf Avg.} \\ \midrule
\multicolumn{13}{l}{\textit{Cross-lingual-transfer (models are fine-tuned on English training data without translation available)}} \\ \midrule
$\text{XLM-R}_{\text{large}}$ & 86.5/75.7 & 68.6/49.0 & 80.4/63.4 & 79.8/61.7 & 82.0/63.9 & 76.7/59.7 & 80.1/64.3 & 74.2/62.8 & 75.9/59.3 & 79.1/59.0 & 59.3/50.0 & 76.6/60.8 \\
\ours{} & 88.9/78.6 & 77.1/60.0 & 83.1/67.2 & 82.6/66.0 & 83.0/65.1 & 77.8/61.8 & 80.8/64.8 & 73.5/62.1 & 77.6/62.0 & 81.8/62.5 & 67.7/58.4 & 79.4/64.4 \\
\midrule
\multicolumn{12}{l}{\textit{Translate-train-all (translation-based augmentation is available for English training data)}} \\ \midrule
FILTER~\citep{DBLP:journals/corr/abs-2009-05166} & 86.4/74.6 & 79.5/60.7 & 83.2/67.0 & 83.0/64.6 & 85.0/67.9 & 83.1/66.6 & 82.8/67.4 & 79.6/73.2 & 80.4/64.4 & 83.8/64.7 & 79.9/77.0 & 82.4/68.0 \\
\ours{} & 88.8/78.1 & 79.7/63.9 & 83.7/68.2 & 83.0/65.7  & 84.7/68.3 & 80.7/64.9 & 82.2/66.6 & 81.9/76.1 & 79.3/65.0 & 82.7/64.5 & 81.3/78.0 & 82.5/69.0 \\
\bottomrule
\end{tabular}
\caption{XQuAD results (F1/EM scores) for each language.}
\label{table:xquad-results}
\end{table*}

\begin{table*}[ht]
\centering
\small
\setlength{\tabcolsep}{1mm}
\begin{tabular}{lcccccccc}
\toprule
{\bf Model} & {\bf en} & {\bf ar} & {\bf de} & {\bf es} & {\bf hi} & {\bf vi} & {\bf zh} & {\bf Avg.} \\ \midrule
\multicolumn{9}{l}{\textit{Cross-lingual-transfer (models are fine-tuned on English training data without translation available)}} \\ \midrule
$\text{XLM-R}_{\text{large}}$ & 83.5/70.6 & 66.6/47.1 & 70.1/54.9 & 74.1/56.6 & 70.6/53.1 & 74.0/52.9 & 62.1/37.0 & 71.6/53.2 \\
\ours{} & 85.2/72.6 & 67.9/47.7 & 72.2/56.8 & 75.5/57.9 & 73.2/55.1 & 75.9/54.7 & 71.1/48.6 & 74.4/56.2 \\
\midrule
\multicolumn{9}{l}{\textit{Translate-train-all (translation-based augmentation is available for English training data)}} \\ \midrule
FILTER~\citep{DBLP:journals/corr/abs-2009-05166} & 84.0/70.8 & 72.1/51.1 & 74.8/60.0 & 78.1/60.1 & 76.0/57.6 & 78.1/57.5 & 70.5/47.0 & 76.2/57.7 \\
\ours{} & 85.3/72.9 & 69.7/50.1 & 72.3/57.3 & 76.3/58.8 & 74.0/56.0 & 76.5/55.9 & 70.8/48.3 & 75.0/57.1 \\
\bottomrule
\end{tabular}
\caption{MLQA results (F1/EM scores) for each language.}
\label{table:mlqa-results}
\end{table*}

\begin{table*}[ht]
\centering
\scriptsize
\setlength{\tabcolsep}{1mm}
\begin{tabular}{lcccccccccc}
\toprule
{\bf Model} & {\bf en} & {\bf ar} & {\bf bn} & {\bf fi} & {\bf id} & {\bf ko} & {\bf ru} & {\bf sw} & {\bf te} & {\bf Avg.} \\ \midrule
\multicolumn{11}{l}{\textit{Cross-lingual-transfer (models are fine-tuned on English training data without translation available)}} \\ \midrule
$\text{XLM-R}_{\text{large}}$ & 71.5/56.8 & 67.6/40.4 & 64.0/47.8 & 70.5/53.2 & 77.4/61.9 & 31.9/10.9 & 67.0/42.1 & 66.1/48.1 & 70.1/43.6 & 65.1/45.0 \\
\ours{} & 75.3/63.6 & 77.4/60.3 & 72.4/58.4 & 75.5/60.2 & 81.5/68.5 & 68.6/58.3 & 71.1/48.8 & 73.3/56.7 & 78.4/60.1 & 74.8/59.4 \\
\midrule
\multicolumn{11}{l}{\textit{Translate-train-all (translation-based augmentation is available for English training data)}} \\ \midrule
FILTER~\citep{DBLP:journals/corr/abs-2009-05166} & 72.4/59.1 & 72.8/50.8 & 70.5/56.6 & 73.3/57.2 & 76.8/59.8 & 33.1/12.3 & 68.9/46.6 & 77.4/65.7 & 69.9/50.4 & 68.3/50.9 \\
\ours{} & 73.8/61.6 & 77.8/60.2 & 73.5/61.1 & 77.0/62.2 & 80.8/68.1 & 66.9/56.5 & 72.1/51.9 & 77.9/65.3 & 77.6/60.7 & 75.3/60.8 \\
\bottomrule
\end{tabular}
\caption{TyDiQA-GolP results (F1/EM scores) for each language.}
\label{table:tydiqa-results}
\end{table*}

\begin{table*}[ht]
\centering
\small
\setlength{\tabcolsep}{1mm}
\begin{tabular}{lccccccccccccccccc}
\toprule
{\bf Model} & {\bf af} & {\bf ar} & {\bf bg} & {\bf de} & {\bf el} & {\bf en} & {\bf es} & {\bf et} & {\bf eu} & {\bf fa} & {\bf fi} & {\bf fr} & {\bf he} & {\bf hi} & {\bf hu} & {\bf id} & {\bf it} \\ \midrule
\multicolumn{18}{l}{\textit{Cross-lingual-transfer (models are fine-tuned on English training data without translation available)}} \\ \midrule
$\text{XLM-R}_{\text{large}}$ & 89.8 & 67.5 & 88.1 & 88.5 & 86.3 & 96.1 & 88.3 & 86.5 & 72.5 & 70.6 & 85.8 & 87.2 & 68.3 & 76.4 & 82.6 & 72.4 & 89.4 \\
\ours{} & 90.4 & 72.8 & 89.0 & 89.4 & 87.0 & 96.1 & 88.8 & 88.1 & 73.1 & 74.7 & 87.2 & 89.5 & 83.5 & 77.7 & 83.6 & 73.2 & 90.5  \\
\midrule
\multicolumn{18}{l}{\textit{Translate-train-all (translation-based augmentation is available for English training data)}} \\ \midrule
FILTER~\citep{DBLP:journals/corr/abs-2009-05166} & 88.7 & 66.1 & 88.5 & 89.2 & 88.3 & 96.0 & 89.1 & 86.3 & 78.0 & 70.8 & 86.1 & 88.9 & 64.9 & 76.7 & 82.6 & 72.6 & 89.8 \\
\ours{} & 90.7 & 74.2 & 89.9 & 90.2 & 87.4 & 96.1 & 90.5 & 88.4 & 75.9 & 74.2 & 87.9 & 90.2 & 85.9 & 79.3 & 83.2 & 73.3 & 91.0 \\ \midrule
{\bf Model} & {\bf ja} & {\bf kk} & {\bf ko} & {\bf mr} & {\bf nl} & {\bf pt} & {\bf ru} & {\bf ta} & {\bf te} & {\bf th} & {\bf tl} & {\bf tr} & {\bf ur} & {\bf vi} & {\bf yo} & {\bf zh} & {\bf Avg.} \\ \midrule
\multicolumn{18}{l}{\textit{Cross-lingual-transfer (models are fine-tuned on English training data without translation available)}} \\ \midrule
$\text{XLM-R}_{\text{large}}$ & 15.9 & 78.1 & 53.9 & 80.8 & 89.5 & 87.6 & 89.5 & 65.2 & 86.6 & 47.2 & 92.2 & 76.3 & 70.3 & 56.8 & 24.6 & 25.7 & 73.8 \\
\ours{} & 62.7 & 78.3 & 55.7 & 82.4 & 90.2 & 88.5 & 90.5 & 63.6 & 88.3 & 61.8 & 94.5 & 76.9 & 72.0 & 57.8 & 24.4 & 69.4 & 78.5  \\
\midrule
\multicolumn{18}{l}{\textit{Fine-tune multilingual model on all target language target language training sets (translate-train-all)}} \\ \midrule
FILTER~\citep{DBLP:journals/corr/abs-2009-05166} & 40.4 & 80.4 & 53.3 & 86.4 & 89.4 & 88.3 & 90.5 & 65.3 & 87.3 & 57.2 & 94.1 & 77.0 & 70.9 & 58.0 & 43.1 & 53.1 & 76.9 \\
\ours{} & 65.3 & 79.8 & 56.0 & 85.5 & 89.7 & 89.3 & 90.8 & 65.7 & 85.5 & 61.4 & 93.8 & 78.3 & 74.0 & 57.5 & 27.9 & 68.8 & 79.3 \\
\bottomrule
\end{tabular}
\caption{POS results (accuracy) for each language.}
\label{table:pos-results}
\end{table*}

\begin{table*}[ht]
\centering
\scriptsize
\setlength{\tabcolsep}{1.3mm}
\begin{tabular}{lcccccccccccccccccccc}
\toprule
{\bf Model} & {\bf en} & {\bf af} & {\bf ar} & {\bf bg} & {\bf bn} & {\bf de} & {\bf el} & {\bf es} & {\bf et} & {\bf eu} & {\bf fa} & {\bf fi} & {\bf fr} & {\bf he} & {\bf hi} & {\bf hu} & {\bf id} & {\bf it} & {\bf ja} & {\bf jv} \\ \midrule
\multicolumn{21}{l}{\textit{Cross-lingual-transfer (models are fine-tuned on English training data without translation available)}} \\ \midrule
$\text{XLM-R}_{\text{large}}$ & 84.7 & 78.9 & 53.0 & 81.4 & 78.8 & 78.8 & 79.5 & 79.6 & 79.1 & 60.9 & 61.9 & 79.2 & 80.5 & 56.8 & 73.0 & 79.8 & 53.0 & 81.3 & 23.2 & 62.5 \\
\ours{} &  85.0 & 80.4 & 59.1 & 84.8 & 79.1 & 80.5 & 82.0 & 78.1 & 81.5 & 64.5 & 65.9 & 82.2 & 81.9 & 62.0 & 75.0 & 82.8 & 55.8 & 83.1 & 30.5 & 65.9 \\
\midrule
\multicolumn{18}{l}{\textit{Translate-train-all (translation-based augmentation is available for English training data)}} \\ \midrule
FILTER~\citep{DBLP:journals/corr/abs-2009-05166} & 83.5 & 80.4 & 60.7 & 83.5 & 78.4 & 80.4 & 80.7 & 74.0 & 81.0 & 66.9 & 71.3 & 80.2 & 79.9 & 57.4 & 74.3 & 82.2 & 54.0 & 81.9 & 24.3 & 63.5 \\
\ours{} & 84.4 & 81.7 & 59.7 & 85.3 & 80.8 & 80.9 & 82.0 & 74.1 & 83.4 & 69.9 & 63.6 & 82.5 & 80.6 & 64.0 & 76.3 & 83.8 & 57.9 & 83.3 & 26.5 & 69.8 \\ \midrule
{\bf Model}  & {\bf ka} & {\bf kk} & {\bf ko} & {\bf ml} & {\bf mr} & {\bf ms} & {\bf my} & {\bf nl} & {\bf pt} & {\bf ru} & {\bf sw} & {\bf ta} & {\bf te} & {\bf th} & {\bf tl} & {\bf tr} & {\bf ur} & {\bf vi} & {\bf yo} & {\bf zh} \\ \midrule
\multicolumn{21}{l}{\textit{Cross-lingual-transfer (models are fine-tuned on English training data without translation available)}} \\ \midrule
$\text{XLM-R}_{\text{large}}$ & 71.6 & 56.2 & 60.0 & 67.8 & 68.1 & 57.1 & 54.3 & 84.0 & 81.9 & 69.1 & 70.5 & 59.5 & 55.8 & 1.3 & 73.2 & 76.1 & 56.4 & 79.4 & 33.6 & 33.1 \\
\ours{} &  76.7 & 57.5 & 65.9 & 68.1 & 73.3 & 67.2 & 63.7 & 85.3 & 84.0 & 73.6 & 70.1 & 66.1 & 60.1 & 1.8 & 76.9 & 83.6 & 76.0 & 80.3 & 44.4 & 38.7 \\
\midrule
\multicolumn{18}{l}{\textit{Translate-train-all (translation-based augmentation is available for English training data)}} \\ \midrule
FILTER~\citep{DBLP:journals/corr/abs-2009-05166} & 71.0 & 51.1 & 63.8 & 70.2 & 69.8 & 69.3 & 59.0 & 84.6 & 82.1 & 71.1 & 70.6 & 64.3 & 58.7 & 2.4 & 74.4 & 83.0 & 73.4 & 75.8 & 42.9 & 35.4 \\
\ours{} & 76.3 & 56.9 & 67.1 & 72.6 & 71.5 & 72.5 & 66.7 & 85.8 & 82.1 & 75.2 & 72.4 & 66.0 & 61.8 & 1.1 & 77.5 & 83.7 & 75.6 & 80.8 & 44.9 & 36.5 \\
\bottomrule
\end{tabular}
\caption{NER results (F1 scores) for each language.}
\label{table:ner-results}
\end{table*}
\end{document}